\def\year{2020}\relax
\documentclass[letterpaper]{article} % DO NOT CHANGE THIS
\usepackage{aaai20}  % DO NOT CHANGE THIS
\usepackage{times}  % DO NOT CHANGE THIS
\usepackage{helvet} % DO NOT CHANGE THIS
\usepackage{courier}  % DO NOT CHANGE THIS
\usepackage[hyphens]{url}  % DO NOT CHANGE THIS
\usepackage{graphicx} % DO NOT CHANGE THIS
\usepackage{graphicx}  % DO NOT CHANGE THIS
\title{Dependable Neural Networks for Safety Critical Tasks}
\author{Molly O'Brien\textsuperscript{\rm 1, 2}, William Goble\textsuperscript{\rm 1}, Greg Hager\textsuperscript{\rm 2}, Julia Bukowski\textsuperscript{\rm 3} \\
\textsuperscript{\rm 1}exida, LLC \\
80 N Main St, Sellersville, PA 18960 \\
\textsuperscript{\rm 2}Department of Computer Science, Johns Hopkins University \\
3400 N Charles St, Baltimore, MD 21218\\ 
\textsuperscript{\rm 3}Department of Electrical and Computer Engineering, Villanova University \\
800 Lancaster Ave, Villanova, PA 19085\\ 
\{mobrien, wgoble\}@exida.com, hager@cs.jhu.edu, julia.bukowski@villanova.edu \\
}
 
\begin{document}

\maketitle

\begin{abstract}
Neural Networks are being integrated into safety critical systems, e.g., perception systems for autonomous vehicles, which require trained networks to perform safely in novel scenarios. It is challenging to verify neural networks because their decisions are not explainable, they cannot be exhaustively tested, and finite test samples cannot capture the variation across all operating conditions. Existing work seeks to train models robust to new scenarios via domain adaptation, style transfer, or few-shot learning. But these techniques fail to predict how a trained model will perform when the operating conditions differ from the testing conditions. We propose a metric, Machine Learning (ML) Dependability, that measures the network's probability of success in specified operating conditions \textit{which need not be the testing conditions}. In addition, we propose the metrics Task Undependability and Harmful Undependability to distinguish network failures by their consequences. We evaluate the performance of a Neural Network agent trained using Reinforcement Learning in a simulated robot manipulation task. Our results demonstrate that we can accurately predict the ML Dependability, Task Undependability, and Harmful Undependability for operating conditions that are significantly different from the testing conditions. Finally, we design a Safety Function, using harmful failures identified during testing, that reduces harmful failures, in one example, by a factor of 700 while maintaining a high probability of success.
\end{abstract}

\section{Introduction}
Neural Networks are being integrated into safety critical, cyber-physical systems, e.g., object detection for autonomous vehicles \cite{grigorescu2019survey}. Relying on learned networks to automate safety critical tasks requires robust network evaluation. Neural Networks (hereafter referred to as \textit{networks}) make decisions that are not explainable. Most networks cannot be exhaustively tested. Recent work shows that network performance can be brittle and change with minimal changes to the input data distributions \cite{recht2018cifar}. It is unclear how to predict a network's performance in an untested scenario; thus, it is unclear how to predict a network's performance in untested operating conditions. 

\subsection{Training Robust Networks}
The Machine Learning (ML) community is actively researching techniques to train models robust to unseen scenarios via domain adaptation, style transfer, or few-shot learning. Prior work has also investigated how to ensure safety \textit{during} network training \cite{turchetta2016safe}, \cite{zhang2019towards}. 

\subsubsection{Domain Adaptation}
Domain adaptation seeks to adjust a trained network to new operating domains. See \cite{csurka2017domain} for a survey of visual domain adaptation techniques. RoyChowdhury et al. propose a method to leverage unlabeled data in a new operating domain to fine-tune a trained network \cite{roychowdhury2019automatic}. RoyChowdhury et al. show an increase in pedestrian detection over baseline for a network trained using sunny images from the Berkely Deep Drive Dataset (BDD100K) \cite{yu2018bdd100k} and adapted to rainy, overcast, snowy day, and night images. Liu et al. address Open Domain Adaptation (generalizing to an unseen target domain) and Compound Domain Adaptation (generalizing to combined target domains) \cite{liu2019compound}. Liu et al. demonstrate results on a compound target of rainy, cloudy and snowy and an open target of overcast images. 
% Porav et al. address the challenge of 
% Don't worry about the weather domain adaptation \cite{porav2019don}. 

\subsubsection{Style Transfer}
In perception, style transfer is used to render images from one domain as if they were from another. Style transfer can be used in safety critical tasks to render a novel scenario in a known style. CycleGANs have achieved impressive results rendering photographs as if they were painted by different artists and transferring the style of similar animals, e.g., rendering a horse as a zebra \cite{Zhu_2017_ICCV}. Gong et al. extend CycleGANs for continuous style generation flowing from one domain to another \cite{gong2019dlow}. Gong et al. demonstrate results transferring styles between object detection datasets. 

\subsubsection{Few-Shot and Zero-Shot Learning}
Few-shot (zero-shot) learning aims to learn a task for given operating conditions with little (no) labeled training data. James et al. use a task embedding to leverage knowledge from previously learned, similar tasks \cite{james2018task} and demonstrate that a robot can learn new tasks with only one real-world demonstration. See \cite{wang2019survey} for a survey of zero-shot learning. 

\subsection{Software Dependability}
Software dependability is defined in \cite{avizienis2004basic} as ``a system's ability to avoid service failures that are more frequent and more severe than acceptable". Initial work improving the Dependability of ML models proposed testing-based approaches to estimate the performance of software when no testing-oracle is available \cite{murphy2008improving}. 

\subsection{Adaptive Network Testing}
Automated test case generation is often necessary in software verification, because most software cannot be tested exhaustively. See \cite{anand2013orchestrated} for an orchestrated survey of automated testing techniques. Adversarial techniques can be used to identify catastrophic failures in networks performing safety critical tasks \cite{uesato2018rigorous}.  Recent work evaluated autonomous vehicles by selecting test scenarios along boundaries where the model’s performance changed quickly \cite{mullins2018adaptive}. Mullins et al. parameterized the testing space by possible variations in the mission and environment and defined test outcomes by mission success or failure and safety success or failure. 

\subsection{Our Contributions}
In ML, network performance is typically measured by the probability of success. We propose that \textit{how} a network fails can be as important as the probability a network will succeed. Specifically, we distinguish between failures that do not violate safety constraints, which we call task failures, and failures that violate safety constraints (whether or not the task is completed), which we call harmful failures. 

In this work, we propose the performance of a network is described by the fraction of successes, task failures, and harmful failures for a given task in specified conditions. To the best of our knowledge, we tackle the previously unaddressed problem: how to evaluate network performance and safety \textit{after} training is complete, when the operating conditions differ from the testing conditions. Following a Notations section, the contents of this paper are as follows: 
\begin{enumerate}
    \item We define ML Dependability\footnote{This is distinct from software Dependability defined in \cite{avizienis2004basic}.} as the probability of completing a task without harm. We define Task Undependability and Harmful Undependability to distinguish failures by the consequences: task failures causing no harm as opposed to harmful failures. 
    \item We develop mathematics to predict the model performance in novel operating conditions by \textit{re-weighting} known test results with knowledge of the novel operating condition probabilities. 
    \item We accurately predict the ML Dependability, Task Undependability, and Harmful Undependability of a network trained to perform a simulated robot manipulation task in novel operating conditions using test results. 
    \item We design a Safety Function to reduce harmful failures in the simulated robot manipulation task under testing conditions. We reduce the harmful failures, in one example, by a factor of 700 while maintaining a high probability of success. 
    \item We discuss how this work can be translated to practical applications and describe directions for future work. 
\end{enumerate}

\section{Notation}
\begin{table}[h]
% \caption{Notation.}\smallskip
\centering
\resizebox{.95\columnwidth}{!}{
\smallskip\begin{tabular}{r l}
$\pi$ & the trained Neural Network \\
$X$ & the set of all possible domain scenarios \\
$x$ & a domain scenario. $x \in X$ \\
$ \pi(x)$ & success indicator for \(\pi\) in scenario \(x\) \\
$ \pi_{avg}(r_d)$ & the average value of \(\pi(x)\) for scenarios \(x\) \\ &  in region \(r_d\)\\
$ \pi^T(x)$ & task failure indicator for \(\pi\) in scenario \(x\) \\
$ \pi^H(x)$ & harmful failure indicator for \(\pi\) in scenario \(x\) \\
$\tau$ & the testing conditions \\
$O$ & the operating conditions \\
\( P_\tau(X), P_O(X)\) & the probability distribution describing \\ & all possible scenarios during testing,\\ & operation (respectively) \\
\( p_\tau(x), p_O(x)\) & the probability of encountering scenario \(x\)\\ &  during testing, operation (respectively) \\
\(D_O(\pi) \) & the dependability of \(\pi\) in conditions \(O\) \\
\(U^T_O(\pi)\) & the task undependability of \(\pi\) in conditions \(O\)\\
\(U^H_O(\pi)\) & the harmful undependability of \(\pi\) \\ & in conditions \(O\) \\ \\
\hline \\
$v$ & the obstacle velocity [inches/second] \\ 
 & in the robot simulation experiments \\
$t$ & the obstacle start time [seconds] in the \\
 & robot simulation experiments \\
$y$ & the robot goal position [inches] in the \\
 & robot simulation experiments \\
\end{tabular}
}
\label{table2}
\end{table}

\section{Methods}
\subsection{Machine Learning Dependability}
In this work we evaluate the performance of a trained, deterministic neural network, $\pi$, performing a safety critical task. A \textbf{domain scenario}, $x$,  is defined as one set of environment conditions and goals for the network. A network may be used iteratively within one scenario, e.g., a controller moving a robot incrementally towards a goal, or used once, e.g., a classifier labelling a sensor reading as valid or faulty. For each scenario, the network attempts to complete a task without causing harm. The outcome of deploying a network in a scenario is the observed \textbf{behavior mode}. We define three behavior modes: success, task failure, and harmful failure. A network is successful if it accomplished the task without causing harm. A task failure occurs when the network failed to complete the specified task but did not cause harm. Any scenario where the network caused harm is labeled a harmful failure, whether or not the task was completed. The \textbf{domain space}, $X$, of a network describes the set of possible domain scenarios. A fully-observed domain includes all variables in the environment and system which impact the outcome of the network. A partially-observed domain includes a subset of the full domain. The \textbf{input space} of a network is defined as the information the network observes. When a network is deployed iteratively, it may observe many inputs for one scenario. The input may include components of the domain space, but need not include the entire domain space. Domain spaces may be numerical or categorical. Note that for a fully-observed numerical domain, one domain scenario $x$ maps to exactly one behavior mode\footnote{\(\pi(x)\) can map to multiple values if  \(x\) does not fully describe the variables that impact the success of the model, i.e., the domain is partially-observed. We define the domain space for modalities like images or speech as partially observed, because many different pixel-values or spectrographs can represent a specified label (a tree in the rain, a man saying ``hello world"). When \(\pi(x)\) cannot be modeled as a constant value, it may be modeled as a distribution. Extending this work to partially-observed domains is an important challenge we hope to address in future work.}. 

We indicate the success of running network \(\pi\) in scenario \(x\) as \(\pi(x)\). \(\pi(x) = 1\) when the model is successful in scenario \(x\); \(\pi(x) = 0\) when the model has a task or harmful failure. \(\pi\) is tested with \(N\) sampled scenarios \(\{x_n\}_{n=1}^N\), \(x_n \sim P_\tau(X)\)  where \(p_\tau(x_n)\) describes the probability of encountering scenario \(x_n\) during testing. We define \textbf{Machine Learning Dependability} as the probability that a model will succeed when operated under specified conditions. We aim to estimate \(D_O(\pi)\): the ML Dependability of model \(\pi\) deployed under the operating conditions described by \(P_O(X)\), where \(P_O(X) \neq P_\tau(X)\)\footnote{The ML Dependability of \(\pi\) under testing conditions, \(D_\tau(\pi)\), is equal to the network accuracy or the fraction of successful tests: \( D_\tau(\pi)=\sum_{i=1}^N{\frac{\pi(x_i)}{N}} \). Likewise, \( U^T_\tau(\pi)=\sum_{i=1}^N{\frac{\pi^T(x_i)}{N}} \) and \( U^H_\tau(\pi)=\sum_{i=1}^N{\frac{\pi^H(x_i)}{N}} \).}.

% In general, the Testing Domain and the Operating Domain need not describe the same set of possible scenarios \(X\). In this analysis we assume \(X\) is the same in testing and operation, but the likelihood of observing a given scenario \(x\) need not be the same during testing and operation. Since the domain is the same in testing and operation we use the terms 'Testing Conditions' and 'Operating Conditions' for clarity. But in general, the 'Testing Domain' and 'Operating Domain' will be used. 

For this analysis, it is assumed that the domain space is numerical and fully observed, that \(P_\tau(X)\) and \(P_O(X)\) are known, and that while \(P_O(X) \neq P_\tau(X)\), both distributions have the same domain space \(X\).

\subsection{Derivation}
\subsubsection{Discrete-Bounded Domain Space}
To begin, we assume \(X\) is discrete with finite \(D\) possible values, \( X = \{x_d\}_{d=1}^D \). The probability distribution describing scenarios during testing is:
\begin{equation}
 P_\tau(X) = \{p_\tau(x_1), ... , p_\tau(x_D)\} = \{p_\tau(x_d)\}_{d=1}^D  
\end{equation}
The probability distribution describing scenarios during operation is:
\begin{equation}
    P_O(X) = \{p_O(x_d)\}_{d=1}^D
\end{equation}
Note that \(P_\tau(X)\) and \(P_O(X)\) can be estimated without testing or operating the network. \textit{As a motivating example, imagine a perception network for an autonomous vehicle. The perception network is trained and tested in Palo Alto but will operate in Seattle. Information like weather patterns can be used to estimate the probability of different scenarios during testing and operation without recording or labelling data in the testing or operating conditions.}

The ML Dependability of network \(\pi\) operating in conditions \(O\) is defined as the probability that model \(\pi\) succeeds when deployed in a scenario \(x\) randomly sampled from the operating conditions \(x \sim P_O(X)\). This is computed as the expected value of \(\pi(x)\), for \(x \sim P_O(X)\).
\begin{equation}
    D_O(\pi) = E[\pi(x)], \quad x \sim P_O(X)
\end{equation}
\begin{equation}
    D_O(\pi) = \sum_{d=1}^D {p_O(x_d)*\pi(x_d)}
\end{equation}
\(P_O(X)\) is known. \(\pi(x_d)\) must be evaluated via testing.
The reader is reminded that the network is fixed and it is assumed the domain space is numerical and fully observed, so \(\pi(x_d)\) is 1 or 0 for a unique \(x_d\). If the domain space of the network is truly discrete and \(D < \infty \), then the network can be exhaustively tested with \(D\) tests. (Note, if \(D\) is finite but large it may be infeasible to exhaustively test the network. This case may be treated as discrete-unbounded.) In most applications, the domain space is discrete-unbounded or continuous so the network cannot be tested exhaustively.  

\subsubsection{Discrete-Unbounded or Continuous Domain Space}
We approximate discrete-unbounded or continuous domain spaces as discrete-bounded by partitioning \(X\) into \(D\) partitions, with \(D < \infty\). Let the \(d^{th}\) partition be defined as the contiguous region \(r_d\) of \(X\), such that \(\dot\cup_{d = 1}^D{r_d}=X\). The reader is reminded that \(N\) test scenarios are drawn from \(P_\tau(X)\) as \(\{x_n\}_{n=1}^N\). \(N_d\) scenarios lie in each partition where \(\{x_i^d\}_{i=1}^{N_d}\) denotes the scenarios in partition \(d\).
We require the partitions are defined so that at least one test scenario lies within each partition, \(N_d > 0,  \forall d \in [0, D]\).
\(P_O(X)\) is equivalently described by:
\begin{equation}
    P_O(X) = \{p_O(r_d)\}_{d=1}^D
\end{equation}
where \(p_O(r_d)\) is computed as:
\(p_O(r_d) = \sum_{x_n \in r_d}{p_O(x_n)} \) for discrete-unbounded domains, or 
\(p_O(r_d) = \int_{r_d}{p_O(x)dx}\) for continuous domains\footnote{Note, \(x\) is not required to be one dimensional.}. $\pi_{avg}(r_d)$ can be estimated as:
% \footnote{This is an estimation because it assumes that each scenario \(x \in r_d\) is equally likely. The ML Dependability within the partition is computed exactly as \(D_{r_d}(\pi) = \sum_{x_d \in r_d}{p_O(x_d)*\pi(x_d)}\) for discrete-unbounded domains or \(D_{rd}(\pi) = \int_{r_d}{p_O(x)\pi(x)dx}\) for continuous domain spaces. Note both of these expressions require infinite test scenarios to compute.}
\begin{equation}
\pi_{avg}(r_d) \approx \frac{\sum_{i=1}^{N_d}{\pi(x_i^d)}}{N_d}
\end{equation}
The overall ML Dependability can now be approximated as:

\begin{equation}
D_O(\pi) \approx \sum_{d=1}^D {p_O(r_d)*\pi_{avg}(r_d)}
\end{equation}

\begin{equation}
   D_O(\pi) \approx \sum_{d=1}^D {p_O(r_d) * \frac{\sum_{i=1}^{N_d}{\pi(x_i^d)}}{N_d} } 
\end{equation}

\subsubsection{Estimating Undependability}
In a similar manner, we can estimate the undependability of the model \(\pi\) in the operating conditions \(O\). \(\pi^{T}(x)=1\) when the task is not completed but no harm is done, and \(\pi^{T}(x)=0\) otherwise. The \textbf{Task Undependability}, \(U^T_O(\pi)\), is the probability that the model will fail to complete the desired task without causing harm in conditions \(O\). We compute the Task Undependability as:
\begin{equation}
    U^T_O(\pi) = E[\pi^{T}(x)], \quad x \sim P_O(X)
\end{equation}

\begin{equation}
    U^T_O(\pi) \approx \sum_{d=1}^D {p_O(r_d) * \frac{\sum_{i=1}^{N_d}{\pi^{T}(x_i^d)}}{N_d}}
\end{equation} \\ \\
\(\pi^{H}(x)=1\) in the event of a harmful failure, and is zero otherwise.
The \textbf{Harmful Undependability} of the model, \(U^H_O(\pi)\), is the probability that the model will cause harm when operated in conditions \(O\), whether or not the task is completed. The Harmful Undependability is computed as:
\begin{equation}
    U^H_O(\pi) = E[\pi^{H}(x)], \quad x \sim P_O(X)
\end{equation}

\begin{equation}
    U^H_O(\pi) \approx \sum_{d=1}^D {p_O(r_d) * \frac{\sum_{i=1}^{N_d}{\pi^{H}(x_i^d)}}{N_d} } 
\end{equation}
Note that success, task failure, and harmful failure are mutually exclusive, so \(D_O(\pi) + U_O^T(\pi) + U_O^H(\pi) = 1\).

\section{Experiments}
\begin{figure}[t]
\centering
\includegraphics[width=0.9\columnwidth]{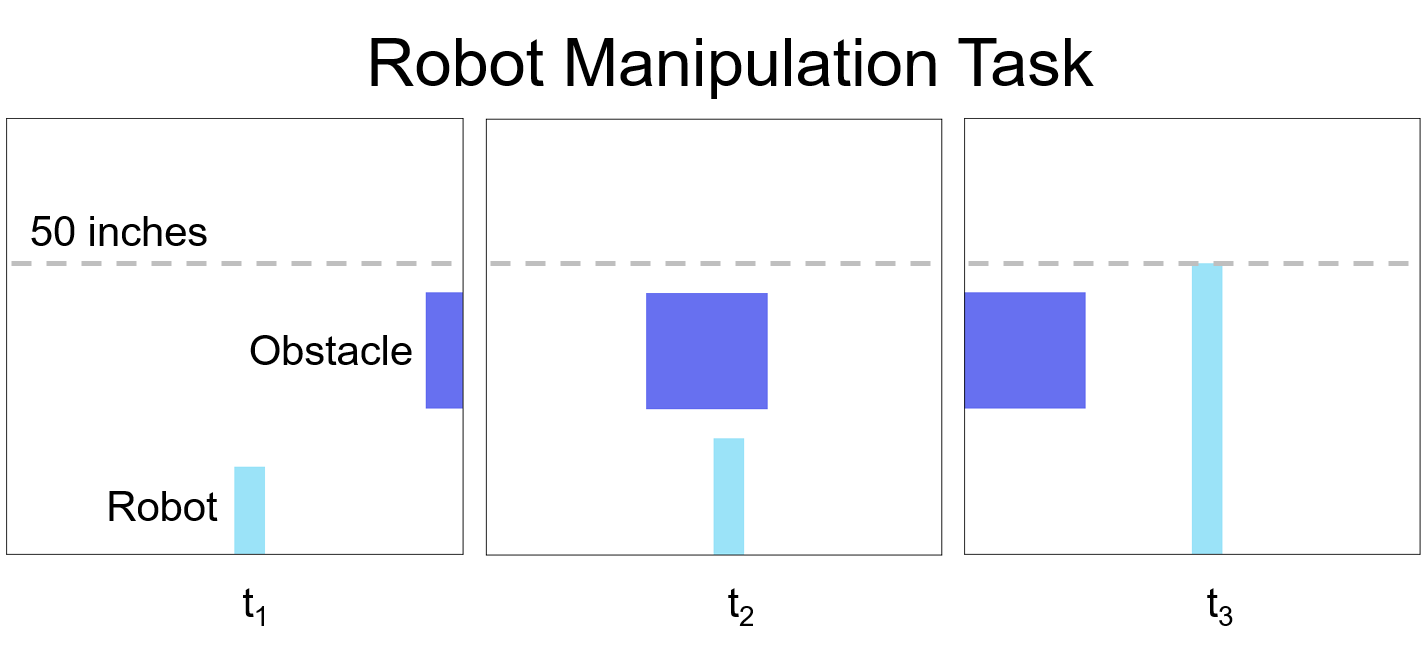} % Reduce the figure size so that it is slightly narrower than the column.
\caption{The simulated robot manipulation task. To succeed, the robot must avoid the obstacle, which moves at a constant velocity \(v\) from right to left, starting at time \(t\), and reach or exceed a goal location, \(y\), between 0 and 50 inches. \(t_1\): the obstacle has started moving. \(t_2\): the robot is avoiding collision with the obstacle. \(t_3\): the robot has successfully reached and/or exceeded its goal position without colliding with the obstacle.}
\label{fig2_robot_env}
\end{figure}

We evaluated the performance of a Neural Network agent trained via Reinforcement Learning to move a simulated robot in the presence of an obstacle that moves at a constant velocity, $v$, starting at time $t$. The obstacle moves from right to left in the scene with its bottom edge 25 inches from the robot base. The robot's task is to reach or exceed a goal position, \(y\), while avoiding the obstacle, see Figure \ref{fig2_robot_env}. The domain space, \(X\), is defined as \(v \in [0, 10]\) inches/second, \(t \in [0, 10]\) seconds, and \(y \in [0, 50]\) inches. The domain space \(X\) is bounded, continuous, and fully observed. The robot starts at 0 inches and is constrained to be within [0, 50] inches\footnote{If the robot tries to move outside this region, the position is clipped. There is no penalty for trying to move outside the valid region.}. The simulations last 100 seconds and the network moves the robot forward 5 inches or back 5 inches every second. The robot moves for the entire 100 second simulation, even after the goal position is reached. A simulation only terminates before 100 seconds if the robot collides with the obstacle.

To succeed, the robot must reach or exceed the goal position before the end of the simulation and avoid the obstacle for the entire simulation. A simulation is a task failure if the robot does not reach the goal position but avoids collision with the obstacle. Any simulation where the robot collides with the obstacle is a harmful failure. In the following results, the behavior modes are denoted with the following colors: success is indicated with green, task failure with blue, and harmful failure with pink. 

The network consists of two linear layers separated by a Rectified Linear Unit (ReLU) and is trained using a modified version of the PyTorch Q-Learning tutorial \cite{pyTorchDQN}. Each second, the network observes the position of the obstacle, the position of the robot, the speed of the obstacle, and the robot goal. Timing information is not input to the network. Zero-mean Gaussian noise with a standard deviation of \(0.1\), \(0.1\), \(0.5\) for \(v, t,\) and  \(y\) respectively is added to the inputs to simulate sensor noise. The reward function for the network was designed so reaching the goal resulted in a reward of 30 points and colliding with the obstacle resulted in a penalty of -50 points. Before reaching the goal position, the network received a small reward of 5 points for moving towards the goal or a penalty of -5 points for moving away from the goal. Before the obstacle had passed the robot, the network received a reward of 2 points for each time step it was below the obstacle and a penalty of -2 points each time step it was in the path of the obstacle.  The point values for reaching the goal (+30 points) and collision (-50 points) were chosen to prioritize safety over task completion. Likewise, the intermediate rewards were chosen so that moving towards the goal ($\pm5$ points) was prioritized above a potential, future collision ($\pm2$ points). 

\subsection{Performance during Testing}
\begin{figure}[h!]
\centering
\includegraphics[width=0.9\columnwidth]{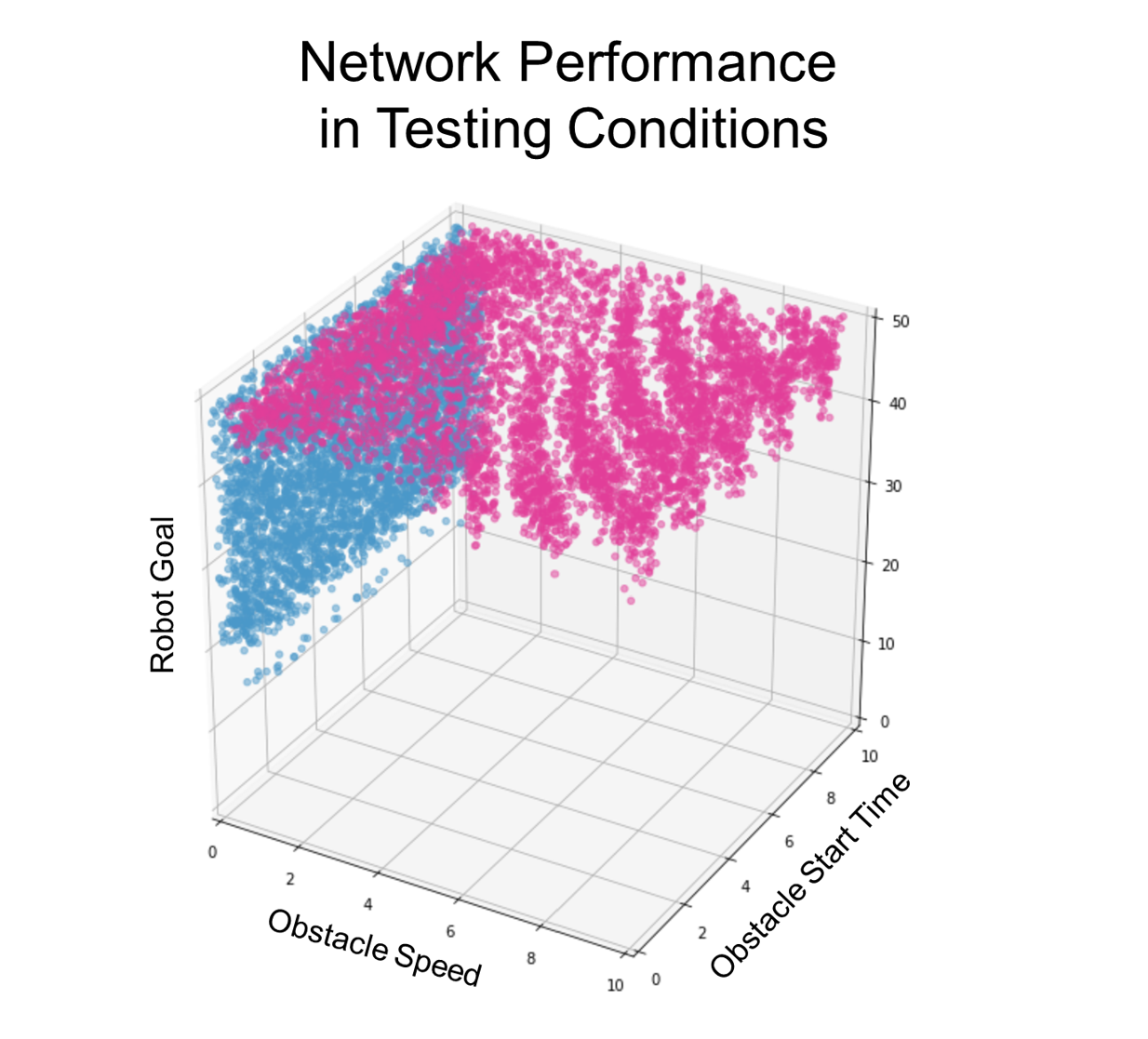} % Reduce the figure size so that it is slightly narrower than the column.
\caption{The observed failures during testing. Blue indicates a task failure. Pink indicates a harmful failure. The task failures (along the left `wall' of the figure) occurred when the obstacle speed was less than or equal to 0.80 inches/second. The harmful failures (along the `ceiling' of the figure) occurred when the robot goal was greater than or equal to 38.47 inches.}
\label{fig_testing_results}
\end{figure}

\(X\) is a bounded, continuous domain space. 
We sample 100,000 test scenarios uniformly from the domain space:
\[P_\tau(X): v \sim U(0, 10), t \sim U(0, 10), y \sim U(0, 50)\]
where \(U(a, b)\) indicates a uniform probability distribution from \(a\) to \(b\). We deployed the trained network in each test scenario to evaluate the network performance. The network had an ML Dependability of \textbf{90.35\%}, a Task Undependability of \textbf{4.18\%}, and a Harmful Undependability of  \textbf{5.47\%}. See Figure \ref{fig_testing_results} for a plot of observed failures by test scenario.

Task failures (shown in blue in Figure \ref{fig_testing_results}) occurred when the obstacle speed was less than or equal to 0.80 inches/second. Inspection revealed that the network learned to wait for the obstacle to pass before moving forward. In many cases the robot moved as far forward as it could, exceeding the input robot goal. When the obstacle moved very slowly, this strategy did not give the network enough time to reach the goal. Harmful failures (shown in pink in Figure \ref{fig_testing_results}) occurred when the robot goal was greater than or equal to 38.47 inches. 

We partition each dimension of the domain space into 10 equal regions to obtain 1,000 voxels in domain space. \(v\) and \(t\) are divided into regions 1 inch/second and 1 second wide (respectively). \(y\) is divided into regions 5 inches wide. We use these voxels to predict the model performance in new operating conditions.

\subsection{Predicting Model Performance in Novel Operating Conditions}
\begin{table}[b]
\caption{Novel Operating Condition Specification. \(\mathcal{N}(\mu, \sigma^2)\) denotes a Gaussian with a mean of \(\mu\) and a standard deviation of \(\sigma\). The sampled scenarios \(x \sim \mathcal{N}(\mu, \sigma^2)\) are clipped to lie within the specified domain \(X\). \( x < X.min \) is set to \(X.min\) and \( x > X.max\) is set to \(X.max\). $t$ is not listed because $t \sim U(0, 10)$ for all conditions.}\smallskip
\centering
\resizebox{.95\columnwidth}{!}{
\smallskip\begin{tabular}{c c c c}
\textbf{Conditions} & \textbf{$v$} &  \textbf{$y$} \\
Testing Conditions & $ \sim U(0, 10)$  & $ \sim U(0, 50)$ \\
Operating Conditions 1 & $ \sim U(0, 10)$  & $ \sim U(0, 30)$ \\
Operating Conditions 2 & $ \sim U(0, 10)$  & $ \sim U(30, 50)$ \\
Operating Conditions 3 & $ \sim \mathcal{N}(3, 2^2)$  & $ \sim U(30, 50)$ \\
Operating Conditions 4 & $ \sim \mathcal{N}(3, 2^2)$  & $ \sim \mathcal{N}(35, 10^2)$ \\
\end{tabular}
}
\label{op_domain_spec_table}
\end{table}
We demonstrate that our method can predict the performance of a network when deployed in novel operating conditions. We define four novel operating conditions in Table \ref{op_domain_spec_table}. The harmful failures in testing occurred for robot goals greater than or equal to 38.47 inches. We selected Operating Conditions 1 to simulate safe conditions: \( y \in [0, 30] \) inches. Operating Conditions 2 simulate dangerous conditions: \( y \in [30, 50]\) inches. We also wanted to select distributions other than uniform (the testing distribution) to make the prediction task more challenging. 
We selected Operating Conditions 3 to introduce a Gaussian domain distribution and focus the obstacle velocity \(v\) towards slower speeds to target the area where task failures occurred. Operating Conditions 4 are the most challenging to predict with Gaussian distributions in $v$ and $y$ focused towards observed task failures and harmful failures.

\begin{figure}[t]
\centering
\includegraphics[width=0.9\columnwidth]{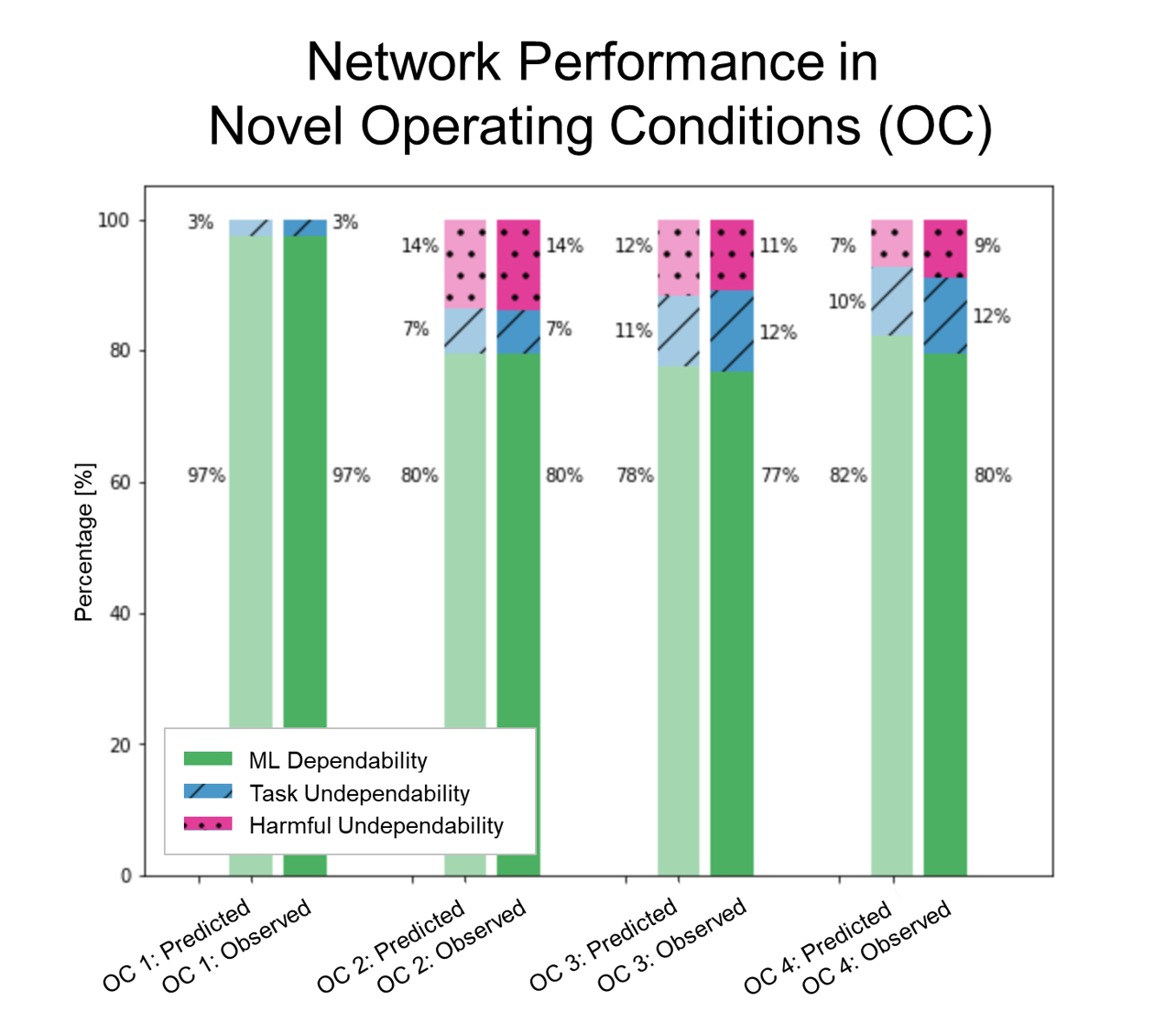} % Reduce the figure size so that it is slightly narrower than the column.
\caption{Predicted and observed performance of the trained network in novel operating conditions. Operating conditions (OC) predicted performance shown left in light colors. Observed performance shown right in bold colors. ML Dependability $D_O(\pi)$ is shown as solid green, Task Undependability $U^T_O(\pi)$ is shown as blue hatched, and Harmful Undependability $U^H_O(\pi)$ is shown as pink dotted bars.}
\label{fig3_op_results}
\end{figure}

We used the partitions defined above to predict the model performance. To confirm our predictions, 100,000 simulations were run for each set of operating conditions. A comparison of our predicted network performance with the observed performance is shown in Figure \ref{fig3_op_results}, above. We accurately predicted the ML Dependability, Task Undependability, and Harmful Undependability within 2\% of observed results.

\begin{figure*}[t!]
\centering
\includegraphics[width=0.8\textwidth]{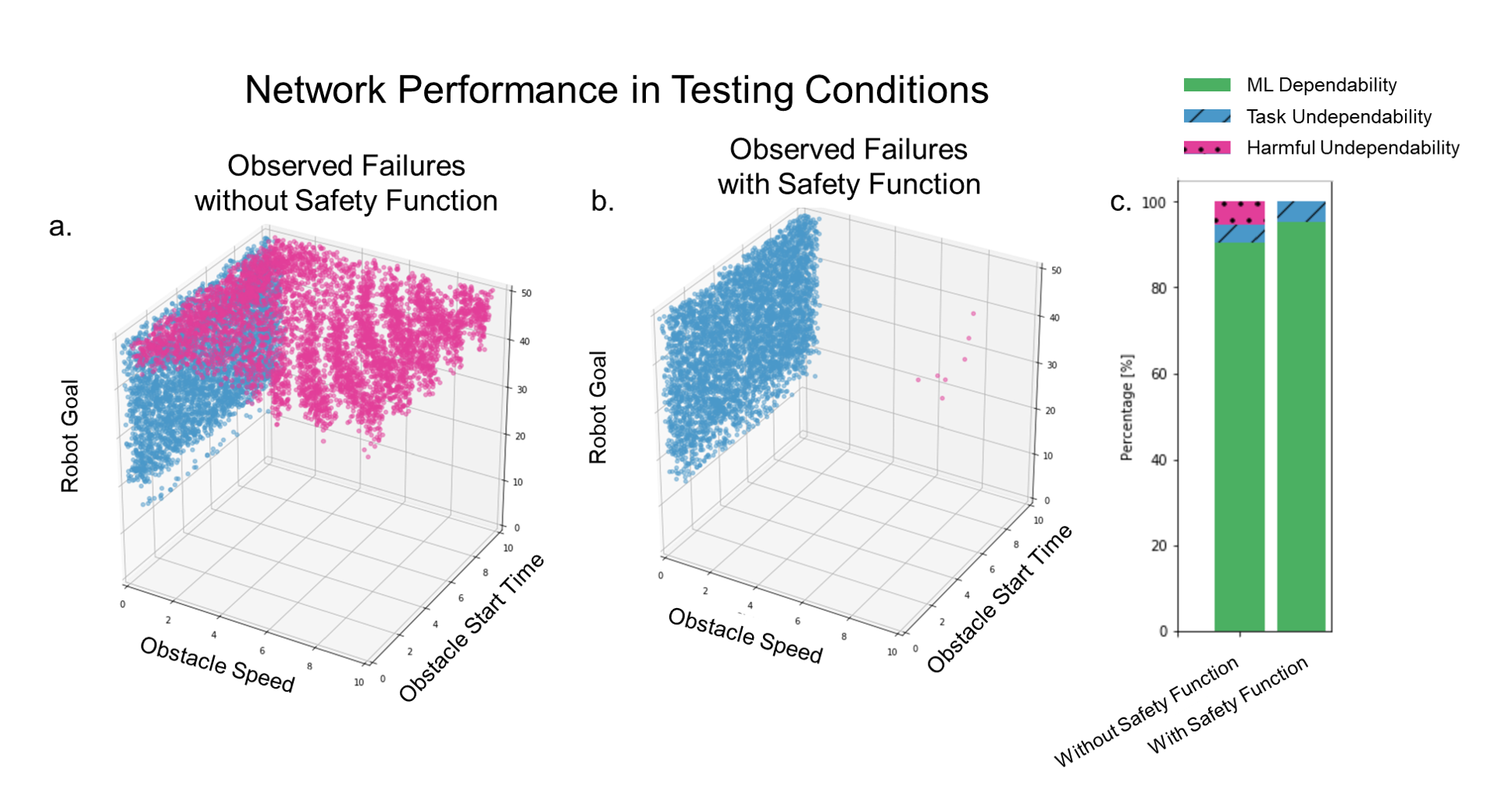} % Reduce the figure size so that it is slightly narrower than the column.
\caption{A comparison of the network performance without the Safety Function and with the Safety Function. Task failures are indicated in blue. Harmful failures are indicated in pink. (a) a reprint of Figure \ref{fig_testing_results} to facilitate comparison. (b) the observed failures in Testing Conditions with the Safety Function. (c) a comparison of the network ML Dependability, Task Undependability, and Harmful Undependability with and without the Safety Function. Note, the Harmful Undependability is reduced from 5.47\% to 0.007\% with the Safety Function.}
\label{fig_testing_wwo_sf}
\end{figure*}

\subsection{Performance with a Safety Function}
Testing revealed that harmful failures only occurred with robot goals greater than or equal to 38.47 inches. We designed a Safety Function to reduce harmful failures by clipping the robot goal input to the network to be between \([0, 38.47-\delta]\) inches. We chose \(\delta=0.5\) inches. The reader is reminded that the network continues to move the robot after the goal position is reached, until the simulation ends at 100 seconds. Clipping the robot goal \textit{input} to the network was intended to make the network behave more conservatively\footnote{This is a similar idea to Control Governors \cite{garone2017reference}.}; it was still possible for the robot to exceed the clipped goal and reach the original goal position. The Safety Function did not change the conditions for success: for a simulation to be successful the robot had to reach the original goal position. 100,000 new test scenarios were sampled from the Testing Conditions and run with the Safety Function. With the Safety Function, the network had a ML Dependability of \textbf{95.19\%}, a Task Undependability of \textbf{4.81\%}, and a Harmful Undependability of \textbf{0.007\%}. Figure \ref{fig_testing_wwo_sf}, above, offers a side-by-side comparison of observed failures and network performance with and without the Safety Function.\\

\section{Discussion}
% Estimating the performance distribution of a network is not trivial -- even for a 2 layer linear network in a numerical domain. 

\subsection{Robot Manipulation Task}
We see in Figure \ref{fig_testing_results} that the network performance varies by region in the domain space. Partitioning the domain space enables these regional variations to emerge when we predict the network performance in novel operating conditions. 

Overall, we accurately predict the performance of the network in novel operating conditions. Across the four proposed operating conditions and three performance metrics, the error between the predicted and observed performance percentage was within \(2\%\). The prediction is poorer for Gaussian domain spaces as compared to uniformly distributed domain spaces. Finer partitioning of the domain space would lead to better predictions and may be necessary as domain space distributions become more complex.

The Safety Function reduced the number of harmful failures by a factor of 700. Surprisingly, even though our Safety Function clipped the input robot goal, it converted many harmful failures into successes. Clipping the robot goal made the network behave more ``patiently", i.e. the network waited for the obstacle to pass before moving as far forward as it could. In general, we expect Safety Functions to reduce the probability of harmful failures, but we do not expect them to increase the probability of success. Our Safety Function was hand-crafted, but in the future, Safety Functions can be learned. It may also be desirable to design or learn different Safety Functions for different operating conditions. Targeted Safety Functions could prove a scalable approach for ensuring safety in dynamic environments, and may be more feasible than retraining the network for different operating conditions.

\subsubsection{Understanding the Network's Behavior}
Both failure modes of the network, task failure and harmful failure, relate to timing. The current time step was not an input to the network; subsequently the network did not learn to make decisions based on timing. The network ML Dependability could be improved in the future by adding a timing input.

Task failures occurred when the obstacle speed was less than or equal to 0.80 inches per second. The network learned to wait for the obstacle to pass the robot before moving past the obstacle, towards the goal. When the obstacle moved slowly this strategy did not give the robot enough time to reach the goal.  But, in these scenarios the network had ample time to reach the robot goal \textit{before} the obstacle passed the robot. Adding a timing input could allow the network to learn more sophisticated timing strategies.

Harmful failures occurred when the robot goal was greater than or equal to 38.47 inches. The network learned an incorrect trade-off between moving towards the goal and avoiding the obstacle. The Safety Function results, see Figure \ref{fig_testing_wwo_sf}, reveal that in most of the scenarios that were harmful failures in testing, the robot had enough time to avoid collision and reach the goal before the end of the simulation. But the strategy learned by the network did not time the robot's approach correctly. Interestingly, the reward function was specifically designed to weight safety over task completion: a collision resulted in a penalty of -50 points whereas reaching the goal resulted in a reward of 30 points. While we do not claim that it would be impossible to craft a reward function to perfectly complete this task without harm, this example illustrates that designing a reward function that appropriately weights task requirements and safety constraints is not trivial. Safety Functions are an explainable alternative to hand crafting reward functions and guarantee a degree of safety for a network.

\subsection{Dependable Networks in Practical Applications}
We make several key assumption in our analysis. The implications of these assumptions determine how this work can be applied in practical applications. We assume that the domain space is numerical. Many applications have numerical domains such as force sensors and distance sensors, e.g. lidar. 

We assume the domain is fully observed. A domain space may be fully observed in a constrained, industrial setting. But as learned networks move into unconstrained, dynamic environments, it is not possible to assume the domain space is fully observed. In partially observed domain spaces, the key change is that we do not assume one scenario \(x\) maps to exactly one output. When we modeled discrete-unbounded and continuous fully observed domain spaces, we modeled the performance of a network in a regions as \( \pi_{avg}(r_d)\). This can be extended in the future to model the distribution of outcomes observed from scenario \(x\) when the domain is only partially observed.  The quality of the performance predictions will vary by how well the partially observed domain describes the full domain. Adequate domain coverage requires expert knowledge. Choosing the dimensions by which we model the domain is an existing challenge and is a direction for further research. Another challenge in modeling practical domain spaces is \textit{the curse of dimensionality}: as the dimension of the domain space grows, the number of partitions or regions can grow prohibitively large. We believe this challenge can be overcome in the future by either selectively choosing the domain to focus on the critical modes of variation for the given application, or leveraging similar scenarios `across' domain variations to limit the effective dimension of the domain space.

We assume \(P_\tau(X)\) and \(P_O(X)\) are known. As stated earlier, \(P_\tau(X)\) and \(P_O(X)\) can be estimated empirically from statistical data or domain knowledge. Lastly, we assume both distributions cover the same domain space \(X\) and that the number of test samples in each partition is greater than zero. This assumption requires some care when designing the partitions.

\subsection{Future Work}
In the future we hope to investigate methods to automatically partition the domain space. We also want to estimate the confidence intervals for predicted ML Dependability, Task Undependability, and Harmful Undependability using the number of samples available in each partition. A rich direction for future research is extending this work to partially-observed domains such as perception. Safety in partially-observed domain spaces is particularly relevant for technology like autonomous vehicles.

\section{Conclusions}
We define and derive the metrics ML Dependability, Task Undependability, and Harmful Undependability to predict a trained network's performance in novel operating conditions. We demonstrate that our metrics can predict the performance of a trained network in novel operating conditions within \(2\%\) of observed performance for a simulated robot manipulation task. We designed a hand-crafted Safety Function to avoid harmful failures identified during testing; the Safety Function was demonstrated to reduce harmful failures by a factor of 700.

\section{Acknowledgments.}
We would like to acknowledge exida, LLC for supporting this work. We would like to thank Mike Medoff, Chris O'Brien,  Andr\'e Ro{\ss}bach, and Austin Reiter for their helpful discussions. 

% References and End of Paper
% These lines must be placed at the end of your paper
\bibliography{ref.bib}
\bibliographystyle{aaai}

\end{document}